\documentclass{article}
\usepackage[dvips]{graphicx}
\usepackage[accepted]{icmlgames2010} 
\usepackage{mlapa}

\icmlgamestitlerunning{Optimizing Selective Search in Chess}

\begin{document} 

\twocolumn[
\icmlgamestitle{Optimizing Selective Search in Chess}

\icmlgamesauthor{Omid David-Tabibi}{mail@omiddavid.com}
\icmlgamesaddress{Department of Computer Science, Bar-Ilan University, Ramat-Gan 52900, Israel}
\icmlgamesauthor{Moshe Koppel}{koppel@cs.biu.ac.il}
\icmlgamesaddress{Department of Computer Science, Bar-Ilan University, Ramat-Gan 52900, Israel}
\icmlgamesauthor{Nathan S.~Netanyahu}{nathan@cs.biu.ac.il}
\icmlgamesaddress{Department of Computer Science, Bar-Ilan University, Ramat-Gan 52900, Israel, and Center
for Automation Research, University of Maryland, College Park, MD 20742}

\vskip 0.3in
]

\begin{abstract} 
In this paper we introduce a novel method for automatically tuning the search parameters of a chess program using genetic algorithms. Our results show that a large set of parameter values can be learned automatically, such that the resulting performance is comparable with that of manually tuned parameters of top tournament-playing chess programs.
\end{abstract}

\section{Introduction}

Until the mid-1970s most chess programs attempted to perform search by mimicking the way humans think, i.e., by generating ``plausible'' moves. By using extensive chess knowledge, these programs selected at each node a few moves which they considered plausible, thereby pruning large parts of the search tree. However, as soon as brute-force search programs like \textsc{Tech} \cite{gillogly72} and \textsc{Chess 4.x} \cite{slate77}
managed to reach depths of 5 plies and more, plausible move generating programs frequently lost to these brute-force searchers due to their significant tactical weaknesses. Brute-force searchers rapidly dominated the computer chess field.

The introduction of null-move pruning \cite{beal89,donninger93} in the early 1990s marked the end of an era, as far as the domination of brute-force programs in computer chess is concerned. Unlike other forward-pruning methods which had great tactical weaknesses, null-move pruning enabled programs to search more deeply with minor tactical risks. Forward-pruning programs frequently outsearched brute-force searchers, and started their own reign which has continued ever since; they have won all World Computer Chess Championships since 1992. \textsc{Deep Blue} \cite{hammilton97,hsu99} was probably the last brute-force searcher.

Nowadays, top tournament-playing programs use a range of methods for adding selectivity to their search. The most popular methods include null-move pruning, futility pruning \cite{heinz98a}, multi-cut pruning \cite{bjornsson98,bjornsson01}, and selective extensions \cite{anantharaman91,beal95}. For each of these methods, a wide range of parameter values can be set. For example, different reduction values can be used for null-move pruning, various thresholds can be used for futility pruning, etc.

For each chess program, the parameter values for various selective search methods are manually tuned through years of experiments and manual optimizations. In this paper we introduce a novel method for automatically tuning the search parameters of a chess program using genetic algorithms (GA).

In the following section, we review briefly the main methods that have been used for selective search. For each of these methods, we enumerate the parameters that need to be optimized. Section 3 provides a review of past attempts at automatic learning of various parameters in chess. In Section 4 we present our automatic method of optimizing the parameters in question, which is based on the use of genetic algorithms, and in Section 5 we provide experimental results. Section 6 contains concluding remarks.

\section{Selective Search in Chess}

In this section we review several popular methods for selective search. All these methods work within the alphabeta/PVS framework and introduce selectivity in various forms. A simple alphabeta search requires the search tree to be developed to a fixed depth in each iteration. Forward pruning methods, such as null-move pruning, futility pruning, and multi-cut pruning, enable the program to prune some parts of the tree at an earlier stage, and devote the time gained to other, more promising parts of the search tree.

Selective extensions, on the other hand, extend certain parts of the tree to be searched deeper, due to tactical considerations associated with a position in question. The following subsections briefly cover each of these pruning and extension methods, and specify which parameters should be tuned for each method.

\subsection{Null-Move Pruning}


Null-move pruning \cite{beal89,david-tabibi08b,donninger93} is based on the assumption that ``doing nothing'' in every chess position (i.e., doing a null-move) is not the best choice even if it were a legal option. In other words, the best move in any position has to be better than the null-move. This assumption enables the program to establish a lower bound $\alpha$ on the position by conducting a \emph{null-move search}. The idea is to make a null-move, i.e., merely swap the side whose turn it is to move. (Note that this cannot be done in positions where the side to move is in check, since the resulting position would be illegal. Also, two null-moves in a row are forbidden, since they result in nothing.) A regular search is then conducted with reduced depth $R$. The returned value of this search can be treated as a lower bound on the position's strength, since the value of the best (legal) move has to be better than that obtained from the null-move search. In a negamax framework, if the returned value is greater than or equal to the current upper bound (i.e., $value \ge \beta$), it results in a cutoff (fail-high). Otherwise, if the value is greater than the current lower bound (i.e., $\alpha < value \le \beta$), we define a narrower search window, as the returned value becomes the new lower bound. If the value is smaller than the current lower bound, it does not contribute to the search in any way. The main benefit of the null-move concept is the pruning obtained due to the cutoffs, which take place whenever the returned value of the null-move search is greater than the current upper bound. Thus, the best way to apply null-move pruning is by conducting a minimal-window null-move search around the current upper bound $\beta$, since such a search will require a reduced search effort to determine if a cutoff takes place.

Donninger \yrcite{donninger93} was the first to suggest an adaptive rather than a fixed value for $R$. Experiments conducted by Heinz in his article on adaptive null-move pruning \yrcite{heinz99} showed that, indeed, an adaptive rather than a fixed value could be selected for the reduction factor. By using $R = 3$ in upper parts of the search tree and $R = 2$ in its lower parts (close to the leaves) pruning can be achieved at a smaller cost (as null-move searches will be shallower in comparison to using a fixed reduction value of $R = 2$) while maintaining the overall tactical strength. An in-depth review of null-move pruning and our \emph{extended null-move reductions} improvement can be found in \cite{david-tabibi08b}.

Over the years many variations of null-move pruning have been suggested, but the set of key parameters to be determined has remained the same. These parameters are: (1) the reduction value $R$, (2) the Boolean adaptivity variable, and (3) the adaptivity depth for which the decremented value of $R$ is applied.

\subsection{Futility Pruning}

Futility pruning and extended futility pruning \cite{heinz98a} suggest pruning nodes near a leaf where the sum of the current static evaluation value and some threshold (e.g., the value of a knight) is smaller than $\alpha$. In these positions, assuming that the value gained in the remaining moves until reaching the leaf is not greater than the threshold, it is safe to assume that the position is ``weak enough'', i.e., that it is worth pruning (as its score will not be greater than $\alpha$). Naturally, the larger the threshold, the safer it is to apply futility pruning, although fewer nodes will be pruned.

The main parameters to be set for futility pruning are: (1) the futility depth and (2) the futility thresholds for various depths (usually up to a depth of 3 plies).

\subsection{Multi-Cut Pruning}

Bj\"ornsson and Marsland's multi-cut pruning \yrcite{bjornsson98,bjornsson01} suggests searching the moves at a given position to a shallower depth first, such that if several of them result in a cutoff, the current node is pruned without conducting a full depth search. The idea is that if there are several moves that produce a cutoff at a shallower depth, there is a high likelihood that at least one of them will produce a cutoff if searched to a full depth. In order to apply multi-cut pruning only to potentially promising nodes, it is applied only to cut-nodes (i.e., nodes at which a cutoff has occurred previously, according to a hash table indication).


The primary parameters that should be set in multi-cut pruning are: (1) the depth reduction value, (2) the depth for which multi-cut is applied, (3) the number of moves to search, and (4) the number of cutoffs to require.

\subsection{Selective Extensions}

Selective extensions \cite{anantharaman91,beal95} are used for extending potentially critical moves to be searched deeper. The following is a list of major extensions used in most programs:

\noindent
\textbf{Check extension:} Extend the move if it checks the opponent's king.\\
\textbf{One-reply extension:} Extend the move if it is the only legal move.\\
\textbf{Recapture extension:} Extend the move if it is a recapture of a piece captured by the opponent (such moves are usually forced).\\
\textbf{Passed pawn extension:} Extend the move if it involves moving a passed pawn (usually to 7th rank).\\
\textbf{Mate threat extension:} Extend the move if the null-move search returns a mate score (the idea is that if doing a null-move results in being checkmated, a potential danger lies at the horizon, so we extend the search to find the threat).

For each of the above extensions, fractional extensions have been widely employed. These are implemented usually by defining one ply to be a number greater than one (e.g., 1 ply = 4 units), such that several fractional extensions along a line (i.e., a series of moves from the root to a leaf) cause a full ply extension. For example, if a certain extension is defined as half a ply, two such extensions must occur along a line in order to result in an actual full ply extension. For each extension type, a value is defined (e.g., assuming that 1 ply = 4 units, an extension has a value between 0 to 4).

From this brief overview of selective search, there are a  number of parameters for each method which have to be set and tuned. Currently, top tournament-playing programs use manually tuned values which take years of trial and improvement to fine tune. In the next section we review the limited success of past attempts at automatic learning of the values of these search parameters, and in Section 4 we present our GA-based method for doing so.

\section{Automatic Tuning of Search Parameters}

The selective search methods covered in the previous section are employed by most of the current top tournament-playing chess programs. They use manually tuned parameter values that were arrived at after years of experiments and manual optimizations.

Past attempts at automatic optimization of search parameters have resulted in limited success. Moriarty and Miikkulainen \yrcite{moriarty94} used neural networks for tuning the search parameters of an Othello program, but as they mention in their paper, their method is not easily applicable to more complex games such as chess. Temporal difference learning has been successfully applied in backgammon and checkers \cite{schaeffer01,tesauro92}. Although the latter has also been applied to chess \cite{baxter00}, the results show that after three days of learning, the playing strength of the program was only 2150 Elo, which is a very low rating for a chess program. Block \emph{et al.}~\yrcite{block08} reported that using reinforcement learning, their chess program achieves a playing strength of only 2016 Elo. Veness \emph{et al.}'s \yrcite{veness09} work on bootstrapping from game tree search improved upon previous work, but their resulting chess program reached a performance of between 2154 to 2338 Elo, which is still considered a very low rating for a chess program. Kocsis and Szepesv\'ari's \yrcite{kocsis06} work on universal parameter optimization in games based on SPSA does not provide any implementation for chess.

Bj\"ornsson and Marsland \yrcite{bjornsson02} presented a method for automatically tuning search extensions in chess. Given a set of test positions (for which the correct move is predetermined) and a set of parameters to be optimized (in their case, four extension parameters), they tune the values of the parameters using gradient-descent optimization. Their program processes all the positions and records, for each position, the number of nodes visited before the solution is found. The goal is to minimize the total node count over all the positions. In each iteration of the optimization process, their method modifies each of the extension parameters by a small value, and records the total node count over all the positions. Thus, given $N$ parameters to optimize (e.g., $N = 4$), their method processes in each iteration all the positions $N$ times. The parameter values are updated after each iteration, so as to minimize the total node count. Bj\"ornsson and Marsland applied their method for tuning the parameter values of the four search extensions: check, passed pawn, recapture, and one-reply extensions. Their results showed that their method optimizes fractional ply values for the above parameters, as the total node count for solving the test set is decreased. 

Despite the success of this gradient-descent method for tuning the parameter values of the above four search extensions, it is difficult to use it efficiently for optimizing a considerably larger set of parameters, which consists of all the selective search parameters mentioned in the previous section. This difficulty is due to the fact that unlike the optimization of search extensions for which the parameter values are mostly independent, other search methods (e.g., multi-cut pruning) are prone to a high interdependency between the parameter values, resulting in multiple local maxima in the search space, in which case it is more difficult to apply gradient-descent optimization.

In the next section we present our method for automatically tuning all the search parameters mentioned in the previous section by using genetic algorithms.

\section{Genetic Algorithms for Tuning of Search Parameters}

In David-Tabibi \emph{et al.}~\yrcite{david-tabibi08a,david-tabibi09,david-tabibi10} we showed that genetic algorithms (GA) can be used to efficiently evolve the parameter values of a chess program's evaluation function. Here we present a GA-based method for optimizing a program's search parameters. We first describe how the search parameters are represented as a chromosome, and then discuss the details of the fitness function. 

The parameters of the selective search methods which were covered in Section 2 can be represented as a binary chromosome, where the number of allocated bits for each parameter is based on a reasonable value range of the parameter. Table~\ref{tab:chromosome} presents the chromosome and the range of values for each parameter (see Section 2 for a description of each parameter). Note that for search extensions fractional ply is applied, where 1 ply $=$ 4 units (e.g., an extension value of 2 is equivalent to half a ply, etc.).

\begin{table}[htbp]
\begin{center}
\begin{tabular}{|l||c|c|}
\hline
Parameter & Value range & Bits\\
\hline
\hline
Null-move use & 0--1 & 1\\
\hline
Null-move reduction & 0--7 & 3\\
\hline
Null-move use adaptivity & 0--1 & 1\\
\hline
Null-move adaptivity depth & 0--7 & 3\\
\hline
Futility depth & 0--3 & 2\\
\hline
Futility threshold depth-1 & 0--1023 & 10\\
\hline
Futility threshold depth-2 & 0--1023 & 10\\
\hline
Futility threshold depth-3 & 0--1023 & 10\\
\hline
Multi-cut use & 0--1 & 1\\
\hline
Multi-cut reduction & 0--7 & 3\\
\hline
Multi-cut depth & 0--7 & 3\\
\hline
Multi-cut move num & 0--31 & 5\\
\hline
Multi-cut cut num & 0--7 & 3\\
\hline
Check extension & 0--4 & 3\\
\hline
One-reply extension & 0--4 & 3\\
\hline
Recapture extension & 0--4 & 3\\
\hline
Passed pawn extension & 0--4 & 3\\
\hline
Mate threat extension & 0--4 & 3\\
\hline
\hline
Total chromosome length & & 70\\
\hline
\end{tabular}
\end{center}
\caption[Chromosome representation of search parameters.]{Chromosome representation of 18 search parameters (length: 70 bits).}
\label{tab:chromosome}
\end{table}

For the GA's fitness function we use a similar optimization goal to the one used by Bj\"ornsson and Marsland \yrcite{bjornsson02}, namely the total node count. A set of 879 tactical test positions from the Encyclopedia of Chess Middlegames (ECM) is used for training purposes. Each of these test positions has a predetermined ``correct move'', which the program has to find. In each generation, each organism searches all the 879 test positions and receives a fitness score based on its performance. As noted, instead of using the number of solved positions as a fitness score, we take the number of nodes the organism visits before finding the correct move. We record this parameter for each position and compute the total node count for each organism over the 879 positions. Since the search cannot continue endlessly for each position, a maximum limit of 500,000 nodes per position is imposed. If the organism does not find the correct move when reaching this maximum node count for the position, the search is stopped and the node count for the position is set to 500,000. Naturally, the higher the maximum limit, the larger the number of solved positions. However, more time will be spent on each position and subsequently, the whole evolution process will take more time.

The fitness of the organism will be inversely proportionate to its total node count for all the positions. Using this fitness value rather than the number of solved positions has the benefit of deriving more fitness information per position. Rather than obtaining a 1-bit information for solving the position, a numeric value is obtained which also measures how quickly the position is solved. Thus, the organism is not only ``encouraged'' to solve more positions, it is rewarded for finding quicker solutions for the already solved test positions.

Other than the special fitness function described above, we use a standard GA implementation with Gray coded chromosomes, fitness-proportional selection, uniform crossover, and elitism (the best organism is copied to the next generation). All the organisms are initialized with random values. The following parameters are used for the GA: population size = 10, crossover rate = 0.75, mutation rate = 0.05, number of generations = 50.

The next section contains the experimental results using the GA-based method for optimization of the search parameters.

\section{Experimental Results}

We used the \textsc{Falcon} chess engine in our experiments. \textsc{Falcon} is a grandmaster-level chess program which has successfully participated in three World Computer Chess Championships. \textsc{Falcon} uses \textsc{NegaScout}/PVS search, with null-move pruning, internal iterative deepening, dynamic move ordering (history + killer heuristic), multi-cut pruning, selective extensions (consisting of check, one-reply, mate-threat, recapture, and passed pawn extensions), transposition table, and futility pruning near leaf nodes.

Each organism is a copy of \textsc{Falcon} (i.e., has the same evaluation function, etc.), except that its search parameters, encoded as a 70-bit chromosome (see Table~\ref{tab:chromosome}), are randomly initialized rather than manually tuned.

The results of the evolution show that the total node count for the population average drops from 239 million nodes to 206 million nodes, and the node count for the best organism drops from 226 million nodes to 199 million nodes. The number of solved positions increases from 488 in the first generation to 547 in the 50th generation. For comparison, the total node count for the 879 positions due to  Bj\"ornsson and Marsland's optimization was 229 million nodes, and the number of solved positions was 508 \cite{bjornsson02}. 

To measure the performance of the best evolved organism (we call this organism \textsc{Evol*}), we compared it against the chess program \textsc{Crafty} \cite{hyatt90}. \textsc{Crafty} has successfully participated in numerous World Computer Chess Championships (WCCC), and is a direct descendent of \textsc{Cray Blitz}, the WCCC winner of 1983 and 1986. It is frequently used in the literature as a standard reference.

First, we let \textsc{Evol*}, \textsc{Crafty}, and the original manually tuned \textsc{Falcon} process the ECM test suite with 5 seconds per position. Table~\ref{tab:search-ecm} provides the results. As can be seen, \textsc{Evol*} solves significantly more problems than \textsc{Crafty} and a few more than \textsc{Falcon}. 

\begin{table}[htbp]
\begin{center}
\begin{tabular}{|c|c|c|}
\hline
\textsc{Evol*} & \textsc{Falcon} & \textsc{Crafty}\\
\hline
\hline
652 & 645 & 593\\
\hline
\end{tabular}
\end{center}
\caption[Number of solved ECM positions (search tuning).]{Number of ECM positions solved by each program (time: 5 seconds per position).}
\label{tab:search-ecm}
\end{table}

The superior performance of \textsc{Evol*} on the ECM test set is not surprising, as it was evolved on this training set. Therefore, in order to obtain an unbiased performance comparison, we conducted a series of 300 matches between \textsc{Evol*} and \textsc{Crafty}, and between \textsc{Evol*} and \textsc{Falcon}. In order to measure the rating gain due to evolution, we also conducted 1,000 matches between \textsc{Evol*} and 10 randomly initialized organisms (\textsc{RandOrg}). Table~\ref{tab:search-matches} provides the results. The table also contains the results of 300 matches between \textsc{Falcon} and \textsc{Crafty} as a baseline.

\begin{table}[htbp]

\begin{center}
\begin{tabular}{|l||c|c|c|}
\hline
Match & Result & W\% & RD\\
\hline
\hline
\textsc{Falcon} - \textsc{Crafty} & 173.5 - 126.5 & 57.8\% & +55\\
\hline
\textsc{Evol*} - \textsc{Crafty} & 178.5 - 121.5 & 59.5\% & +67\\
\hline
\textsc{Evol*} - \textsc{Falcon} & 152.5 - 147.5 & 51.1\% & +6\\
\hline
\textsc{Evol*} - \textsc{RandOrg} & 714.0 - 286.0 & 71.4\% & +159\\
\hline
\end{tabular}
\end{center}
\caption[Performance of the evolved search parameters.]{\textsc{Falcon} vs.~\textsc{Crafty}, and \textsc{Evol*} vs.~\textsc{Crafty}, \textsc{Falcon}, and randomly initialized organisms (W\% is the winning percentage, and RD is the Elo rating difference).}
\label{tab:search-matches}

\end{table}

The results of the matches show that the evolved parameters of \textsc{Evol*} perform on par with those of \textsc{Falcon}, which have been manually tuned and refined for the past eight years. Note that the performance of \textsc{Falcon} is by no means a theoretical upper bound for the performance of \textsc{Evol*}, and the fact that the automatically evolved program matches the manually tuned one over many years of world championship level performance, is by itself a clear demonstration of the capabilities achieved due to the automatic evolution of search parameters.

The results further show that \textsc{Evol*} outperforms \textsc{Crafty}, not only in terms of solving more tactical test positions, but more importantly in its overall strength. These results establish that even though the search parameters are evolved from scratch (with randomly initialized chromosomes), the resulting organism outperforms a grandmaster-level chess program.

\section{Conclusions}

In this paper we presented a novel method for automatically tuning the search parameters of a chess program. While past attempts yielded limited success in tuning a small number of search parameters, the method presented here succeeded in evolving a large number of parameters for several search methods, including complicated interdependent parameters of forward pruning search methods.

The search parameters of the \textsc{Falcon} chess engine, which we used for our experiments, have been manually tuned over the past eight years. The fact that GA manages to evolve the search parameters automatically, such that the resulting performance is on par with the highly refined parameters of \textsc{Falcon} is in itself remarkable.

Note that the evolved parameter sets are not necessarily the best parameter sets for every chess program. Undoubtedly, running the evolutionary process mentioned in this paper on each chess program will yield a different set of results which are optimized for the specific chess program. This is due to the fact that the performance of the search component of the program depends on other components as well, most importantly the evaluation function. For example, in a previous paper on \emph{extended null-move pruning} \cite{david-tabibi08b}, we discovered that while the common reduction value for null-move pruning is $R = 2$ or $R = 3$, a more aggressive reduction value of adaptive $R = 3 \sim 4$ performs better for \textsc{Falcon}. It is interesting to note that our GA-based method managed to independently find that these aggressive reduction values work better for \textsc{Falcon}.

\bibliographystyle{mlapa}

\end{document}